\documentclass[a4paper,fleqn]{cas-dc}

\usepackage[authoryear,longnamesfirst]{natbib}
\usepackage{siunitx}

\usepackage{xcolor}
\usepackage[most]{tcolorbox}
\definecolor{lightgray}{gray}{0.9}

\newtcolorbox{highlightboxinternal}{
  colback=lightgray,
  colframe=lightgray,
  boxrule=0pt,
  sharp corners,
  width=\columnwidth,
  left=1em,
  right=1em,
  top=0.5\baselineskip,
  bottom=0.5\baselineskip,
  before skip=0.25cm,
  after skip=0.25cm,
  parbox=false,
  breakable
}

\def\tsc#1{\csdef{#1}{\textsc{\lowercase{#1}}\xspace}}
\tsc{WGM}
\tsc{QE}

\begin{document}
\let\WriteBookmarks\relax
\def\floatpagepagefraction{1}
\def\textpagefraction{.001}

\shorttitle{}    

\shortauthors{P. Pavlík et~al.}  

\title [mode = title]{Assessing the Utility of Volumetric Motion Fields for Radar-based Precipitation Nowcasting with Physics-informed Deep Learning}



%

\author[1,2]{P. Pavlík}[orcid=0000-0002-7468-5503]
\cormark[1]
\cortext[1]{Corresponding author}
\ead{peter.pavlik@kinit.sk}
\credit{Conceptualization, Methodology, Software, Data Curation, Formal analysis, Investigation, Visualization, Writing – original draft}

\affiliation[1]{organization={Kempelen Institute of Intelligent Technologies},
            addressline={Sky Park Offices, Bottova 7939/2A}, 
            city={Bratislava},
            citysep={},
            postcode={811 09},
            country={Slovakia}}

\affiliation[2]{organization={Faculty of Information Technology, Brno University of Technology},
            addressline={Božetěchova 2}, 
            city={Brno},
            citysep={},
            postcode={612 00},
            country={Czech Republic}}

\author[1]{A. {Bou Ezzeddine}}[orcid=0000-0002-3341-6059]
\ead{anna.bou.ezzeddine@kinit.sk}
\credit{Supervision, Writing - Review \& Editing}


\author[1]{V. Rozinajová}[orcid=0000-0003-1302-6261]
\ead{viera.rozinajova@kinit.sk}
\credit{Project administration, Supervision, Resources, Writing - Review \& Editing}


\begin{abstract}
Estimating motion from spatiotemporal geoscientific data is a fundamental component of many environmental modeling and forecasting tasks. In this work, we propose a physics-informed deep learning framework for estimating altitude-wise motion fields directly from volumetric radar reflectivity data. The model utilizes a fully differentiable semi-Lagrangian extrapolation operator to process three-dimensional inputs as independent horizontal slice sequences, enabling efficient inference of horizontal motion across multiple altitude levels. Using a multi-year radar dataset from Central Europe, we evaluate the impact of altitude-wise motion estimation on extrapolation-based precipitation forecasting and conduct a systematic dataset-scale analysis of inter-altitude motion consistency. The results show that the estimated motion fields exhibit strong vertical coherence, with high correlation across altitude levels, which results in limited improvement over traditional two-dimensional approach in this setting. The proposed framework provides a general tool for efficiently analyzing motion structure in volumetric geospatial data. The findings indicate that, in regions dominated by vertically coherent precipitation systems, the added complexity of volumetric motion modeling may offer limited benefit, warranting careful consideration in the design of efficient spatiotemporal advection models.
\end{abstract}





\maketitle

\section{Introduction} \label{sec:intro}

Precipitation nowcasting refers to the task of predicting the intensity and spatial distribution of precipitation (rain, snow, hail, etc.)  over the next few hours with high local accuracy~\cite{WMO17}. These short-term forecasts are crucial for early warning systems, helping to reduce damage and save lives during extreme weather events. As the climate crisis drives greater variability in precipitation patterns~\cite{Zhang2024}, improving nowcasting capabilities is becoming key in making the world's societies and economies more resilient.

Precipitation nowcasting typically relies on observations from ground-based weather radars, which estimate the location and intensity of hydrometeors within their coverage area. In its simplest form, nowcasting is performed by examining a sequence of past radar images, inferring the recent motion, and extrapolating the latest observation forward in time accordingly. This principle forms the foundation of traditional optical-flow-based approaches, which, despite their simplicity, can outperform numerical weather prediction (NWP) models at lead times of up to a few hours~\cite{NOVAK2009328}.

To compensate for the limitations of simple radar extrapolation, traditional nowcasting systems typically include an additional post-processing step that estimates the evolution of precipitation -- its growth, decay, and structural changes. Historically, statistical or heuristic approaches have been leveraged to adjust the extrapolated radar fields according to the trends in recent observations. Examples of traditional algorithms that also predict growth and decay to some extent using statistical methods are S-PROG~\cite{SPROG} with its extensions STEPS~\cite{bowler2006steps} and LINDA~\cite{pulkkinen2021lagrangian}; or ANVIL~\cite{pulkinnen2020anvil}.

In recent years, deep learning models have transformed the landscape of precipitation nowcasting. Early works like \mbox{ConvLSTM}~\cite{shi2015convolutional}, PredRNN~\cite{wang2017predrnn}, or RainNet~\cite{ayzel2020rainnet}, treat nowcasting as a direct sequence-to-sequence prediction problem, taking several past radar frames as input and generating future fields. These models often outperformed classical statistical post-processing, but they struggled to fully capture the physical mechanisms governing precipitation without explicit modeling of the displacement due to advection.

When machine learning-based models are trained to minimize traditional gridpoint-based loss functions, the chaotic nature of the atmosphere gives rise to the so-called double penalty problem: a forecast that accurately captures the intensity, size, and timing of a precipitation feature but not the advection displacement is penalized twice -- once for missing the event at the true location and once for predicting it at the wrong one~\cite{Keil2009}. As a result, models tend to produce mathematically optimal predictions that appear overly smooth and blurry, lacking physical realism. This smoothing effect particularly degrades the representation of extreme events, which are dampened or smeared out, thereby limiting the practical value of such forecasts for early warning applications.

More recent deep learning models explicitly incorporate motion modeling into their architectures. Models such as TrajGRU~\cite{shi2017deep}, NowcastNet~\cite{zhang2023skilful}, or L-CNN~\cite{lcnn} introduce learned motion fields or motion-aware components that separate the tasks of advection and precipitation field evolution. This hybrid paradigm that combines explicit movement modeling with deep learning-based estimation of growth and decay has led to major improvements in spatial accuracy at longer lead times, mitigating the blurring effect of the double penalty problem~\cite{pavlik2025fully}.

Most existing nowcasting approaches operate on two-dimensional radar composites and therefore assume that precipitation advection is sufficiently represented in a vertically collapsed space. This assumption simplifies the problem and is totally adequate in many cases, but it neglects the three-dimensional structure of atmospheric motion, where vertical wind shear and inter-altitude flow differences may influence the horizontal displacement of precipitation features.

Some prior studies have explored the use of three-dimensional radar observations for precipitation nowcasting. Three-dimensional convolutional neural networks have been applied to volumetric radar data and demonstrated improvements in forecast skill, suggesting that vertical structure and height-dependent motion can provide additional predictive information~\cite{KIM2021105774,pavlik2022radar,sun2022three}. In addition to fully data-driven approaches, three-dimensional optical flow methods have also been explored~\cite{sun2022three,chung2025investigating}. The Nowcast3D model successfully employed a novel deep learning architecture to estimate volumetric motion fields from radar data, reporting improved performance across a range of evaluation scenarios~\cite{chen2025nowcast3d}. However, the extent to which explicitly modeled volumetric motion contributes to nowcasting performance, independent of other factors, remains unclear.

In this work, we train a deep learning model to estimate volumetric horizontal motion fields from sequences of radar reflectivity volumes and examine whether explicitly modeling three-dimensional precipitation advection offers practical advantages for precipitation nowcasting. To focus specifically on the role of advection modeling, we exclude precipitation growth and decay processes from the forecast, allowing for a more controlled comparison between two- and three-dimensional motion representations.

\section{Optical Flow and Nowcasting} \label{sec:related_work}

Most traditional precipitation nowcasting methods build on the assumption of persistence -- precipitation in the near future will closely resemble the present state. In an Eulerian framework, this assumption corresponds to simply replicating the most recent observation, but the accuracy of such nowcast quickly drops as precipitation systems move elsewhere.

A more realistic formulation adopts a Lagrangian perspective, in which precipitation features are assumed to persist while being transported by an advection (motion) field. Under this view, nowcasting is reduced to estimating a motion field and advecting the most recent precipitation observation forward in time according to this field. This assumption forms the basis of classical radar extrapolation and optical-flow-based nowcasting methods. Under this view, the temporal evolution of the precipitation field is governed by the advection equation:

\begin{equation} \frac{\partial \Psi }{\partial t}+\nabla \cdot (\Psi \boldsymbol{u})=0 \label{eq:adv_eq} \end{equation}

where $\Psi$ denotes the 2D radar precipitation field and $\boldsymbol{u}$ the horizontal motion field. It is commonly assumed that the flow is incompressible, i.e., $\nabla \cdot \boldsymbol{u} = 0$.

This formulation expresses the Lagrangian persistence assumption -- the change in precipitation intensity along flow trajectories defined by the motion field $\boldsymbol{u}$ is zero. Traditional radar extrapolation and optical-flow-based nowcasting methods rely directly on this assumption by estimating  $\boldsymbol{u}$ from recent observations and advecting the most recent precipitation field forward in time.

In practice, the quality of Lagrangian nowcasts naturally heavily depends on the accuracy of the estimated motion field. Classical methods typically infer the motion field $\boldsymbol{u}$ from sequences of 2D radar composites using optical flow techniques such as the Lucas-Kanade algorithm~\cite{lucas1981iterative}, assuming that precipitation motion can be adequately represented in two dimensions.

An example of an optical flow approach taking into account the full 3D extent of the radar data is provided in~\cite{chung2025investigating}, where the spatiotemporal characteristics of motion fields derived from three-dimensional radar echoes over Taiwan were investigated. In their work, the motion field has no vertical velocity component, but it is estimated independently for each altitude level, thereby allowing the horizontal advection to vary with height. Their analysis shows that this altitude-wise formulation can be beneficial in situations characterized by strong vertical variability of echo motion, particularly during deep convective events, Meiyu front (eastward mesoscale convective system), and typhoons interacting with complex terrain. In these scenarios, significant differences in motion speed and direction were observed between lower and upper levels, leading to reduced inter-altitude correlation and improved representation of echo advection compared to two-dimensional composite-based motion fields. Conversely, the authors report that in stratiform precipitation with limited vertical echo extent, motion fields remain highly correlated across heights, limiting the practical benefit of the volumetric approach.

Moving beyond the altitude-wise separation of horizontal motion, recent deep learning frameworks have attempted to directly model the full three-dimensional evolution of weather systems by coupling physical operators with data-driven representations. A prominent example is the Nowcast3D~\cite{chen2025nowcast3d} model, which treats radar reflectivity evolution not merely as advection, but as a composite of three physical processes: three-dimensional advection, local diffusion, and a source term representing microphysical tendencies. Unlike the altitude-wise motion estimation in~\cite{chung2025investigating}, Nowcast3D estimates a complete 3D velocity field, including the vertical component, directly from volumetric reflectivity. While the authors report superior performance over operational baselines in forecasting extreme convective events, they also acknowledge inherent ambiguity in the approach. As the authors note, the inference of a full 3D motion field solely from reflectivity observations remains an under-constrained problem, particularly in regimes dominated by complex microphysics where phase changes alter reflectivity without corresponding changes in velocity.

The findings of~\cite{chung2025investigating} and~\cite{chen2025nowcast3d} suggest that modeling volumetric flow does not universally translate into improved nowcasting performance, motivating a careful evaluation of when volumetric motion modeling is warranted.

\section{Dataset} \label{sec:dataset}

Many operational nowcasting systems rely on two-dimensional products such as column-maximum reflectivity (CMAX) or constant-altitude plan position indicators (CAPPI). However, volumetric radar data offers additional information about the vertical extent and structure of precipitation echoes that cannot be recovered from 2D composites alone.

\begin{figure}
    \centering
    \includegraphics[width=\columnwidth]{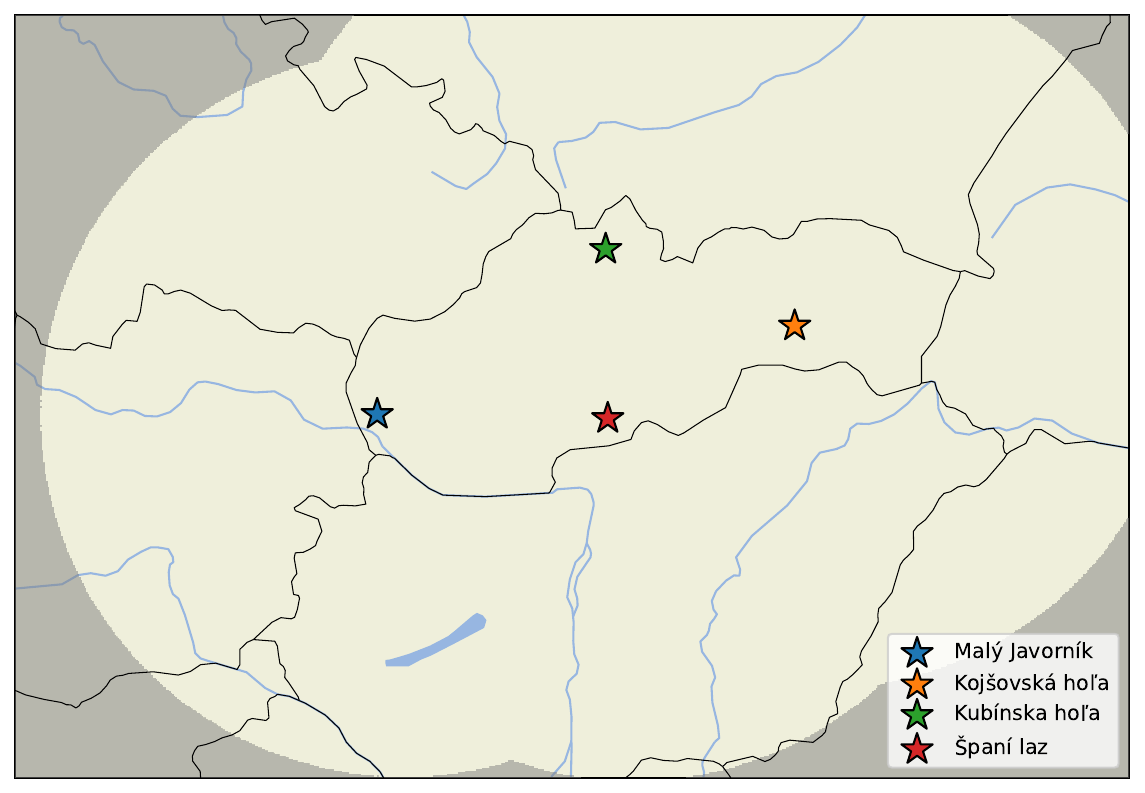}
    \caption{The figure shows the locations of the four ground radar stations comprising the Slovak radar network. The overlapping lightly shaded circular areas show the maximum extent of the radar coverage.}
    \label{fig:extent}
\end{figure}

In this study, we use volumetric radar reflectivity data from the Slovak national weather radar network consisting of four C-band Doppler radar stations (see Fig.~\ref{fig:extent}). Climatological analysis of thunderstorm occurrences in Slovakia from 1984–2023 documents a pronounced seasonal cycle with peak activity in June and July, spatial variability related to topography, and typical frequencies of 16 thunderstorm days per year, consistent with a continental, mid-latitude convective regime~\cite{vido2024thunderstorm}.

The radar network provides volumetric scans at 5-minute intervals, enabling the analysis of precipitation evolution at temporal resolutions suitable for short-term nowcasting. The dataset spans the period of four years from 12 June 2018 to 22 August 2022 and is curated to contain observations from events with significant radar echo measurements. The selected events are approximately uniformly distributed throughout the year, covering a wide range of meteorological conditions. We mapped the radar observations to a Cartesian grid with a horizontal resolution of 1~km $\times$ 1~km, resulting in images of size $517 \times 755$ pixels. Vertically, each radar volume spans 16 altitude levels ranging from 500~m to 8000~m above ground level, capturing both low-level precipitation structures and deeper storm development.

Radar measurements are provided in terms of equivalent reflectivity factor (\si{\dB Z}). For model training and evaluation, reflectivity values were converted to precipitation intensity expressed in \si{\mm\per\hour} using the Marshall-Palmer Z-R reflectivity-rain rate relationship~\cite{marshall-palmer}. This presents a more physically interpretable learning target useful both for model training and evaluation.

\begin{figure*}
    \centering
    \begin{minipage}{0.49\textwidth}
        \centering
        \includegraphics[width=\linewidth]{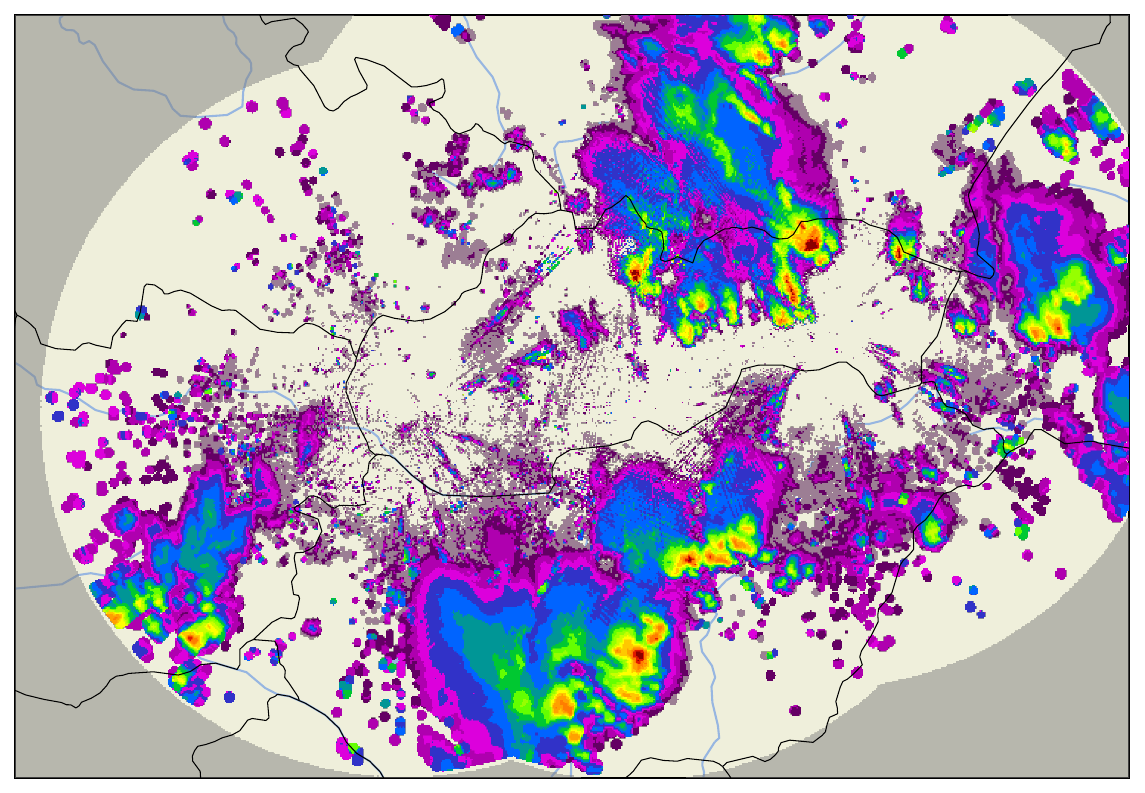}\\
        (a)
    \end{minipage}\hfill
    \begin{minipage}{0.49\textwidth}
        \centering
        \includegraphics[width=\linewidth]{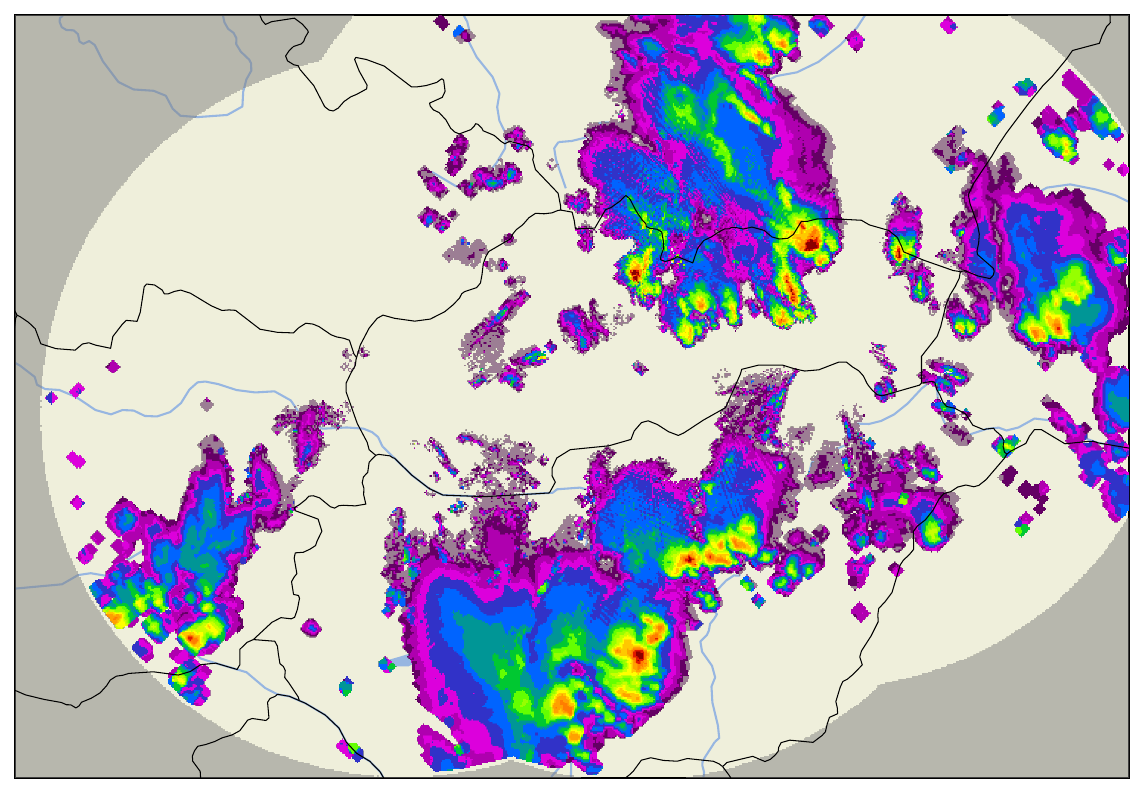}\\
        (b)
    \end{minipage}
    \caption{Example of the radar reflectivity denoising process: (a) raw reflectivity field containing non-meteorological echoes and small-scale noise, and (b) corresponding reflectivity field after polarimetric filtering and morphological denoising. Both of the images are created by displaying the vertical column maximum from the 3D volume for each pixel (CMAX).}
    \label{fig:denoise}
\end{figure*}

In the dataset, non-meteorological echoes and small-scale noise were suppressed using a combination of polarimetric filtering and morphological operations applied to the radar reflectivity fields. The comparison between the raw observation and the result of the denoising process is shown in Figure~\ref{fig:denoise}.

\subsection{Preliminary Data Analysis} \label{sec:PDA}

\begin{figure*}
    \centering
    \includegraphics[width=\textwidth]{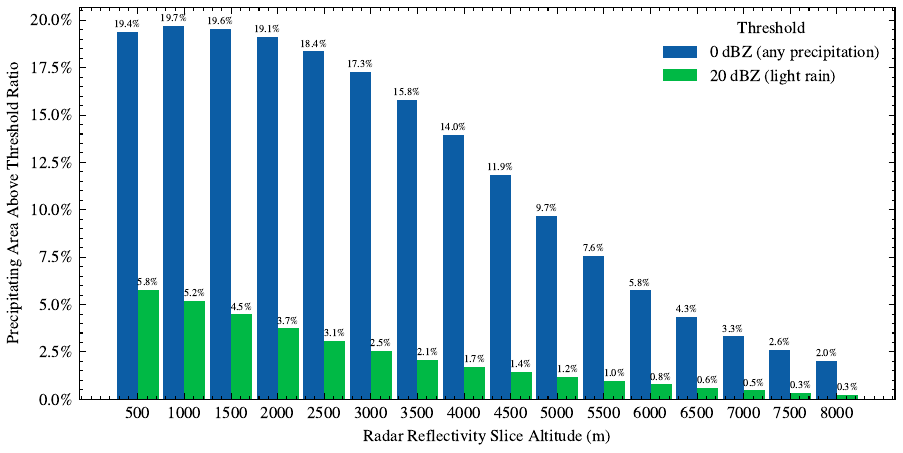}
    \caption{Bar chart showing the mean fraction of grid points with reflectivity exceeding 0 and 20~\si{\dB Z} at each altitude level. A rapid decrease in rainy pixel occurrence with height is observed, reflecting the limited vertical extent of many precipitation events in the dataset.}
    \label{fig:reflectivity_bars}
\end{figure*}

To learn more about the vertical structure of precipitation in the dataset, we compute, for each radar volume sample, the fraction of grid points exceeding specified reflectivity thresholds at each altitude level. We use two reflectivity thresholds: 0~\si{\dB Z} to quantify the overall occurrence of any precipitation echoes and 20~\si{\dB Z} to represent light rain and above. See the calculated ratios in Figure~\ref{fig:reflectivity_bars}.

The fraction of rainy pixels decreases rapidly with altitude for both thresholds. These results indicate that most of precipitation events in the dataset exhibit limited vertical extent, which does not present a promising outlook for the utility of volumetric motion fields in the general case.

\begin{figure}
    \centering
    \includegraphics[width=\columnwidth]{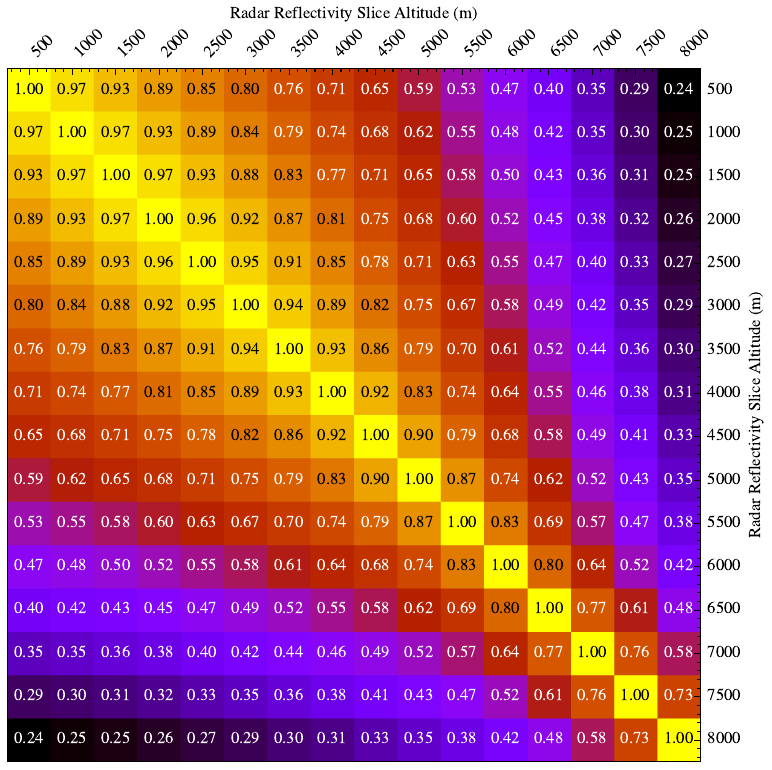}
    \caption{Correlation matrix showing the mean pixel-wise Pearson correlation coefficients between horizontally co-located radar reflectivity slices at different altitude levels from 500 to 8000 m above ground level. Only samples where non-zero reflectivity was observed at each altitude were included in the calculation.}
    \label{fig:reflectivity_corr}
\end{figure}

Additionally, we analyze the degree of similarity between precipitation structures observed at different altitude levels. We can assume that strongly correlated reflectivity patterns across altitudes are likely to give rise to similar motion estimates, thereby reducing the potential benefit of volumetric motion fields. We compute the Pearson correlation coefficient between all pairs of altitude-specific reflectivity volume slices. Correlations are evaluated on a subset of samples in which non-zero reflectivity is present at all altitude levels. The resulting correlation matrix is shown in Fig.~\ref{fig:reflectivity_corr}.

The correlation matrix reveals a high degree of vertical coherence in radar reflectivity structures across the majority of altitude levels, not only between adjacent layers, but also across several kilometers of vertical separation and only gradually decorrelate toward the highest altitudes. As similar spatial patterns across heights are likely to produce highly similar motion estimates, the potential benefit of estimating independent motion fields for each altitude layer is expected to be limited in the general case. However, volumetric motion estimation might still prove beneficial for certain events with lower inter-altitude correlation and should be investigated. Therefore, we proceed with a systematic evaluation of volumetric motion field estimation, with the aim of identifying scenarios in which height-resolved advection provides measurable improvements over two-dimensional approaches.

\section{3DMF-U-Net: Volumetric Motion Field U-Net} \label{sec:methods}

We adopt an approach in which the volumetric radar observations are decomposed into a set of two-dimensional horizontal altitude slices, and a separate horizontal-only motion field is estimated for each slice using a 3D U-Net architecture (meaning a U-Net consisting of 3D convolutional layers) with time treated as an explicit third dimension. The resulting altitude-wise horizontal motion fields are subsequently stacked to produce a pseudo-volumetric motion field without an explicit vertical velocity component.

We took inspiration from the approach employed in \cite{chung2025investigating}, where horizontal motion fields were computed independently for each two-dimensional horizontal slice of the volumetric radar volume sequence. The model predicts a single time-invariant motion field per slice, intended to represent the dominant horizontal advection governing near-future precipitation evolution. The horizontal slicing simplifies the task, but has a few benefits:

\begin{itemize}
    \item Possibility of treating time as a third dimension, enabling more effective spatiotemporal feature extraction~\cite{nagasato2021extension,ye2022msstnet,pavlik2025echo} and improving the performance of the model, without requiring 4D convolutional neural layers.
    \item An identical neural network architecture can be used both for the 2D CMAX benchmark model and for the volumetric case (applied to each altitude slice), ensuring a fair and controlled comparison between 2D and volumetric motion estimation.
    \item Improved interpretability, as motion fields and the accordingly extrapolated reflectivity fields can be directly analyzed and compared across individual altitude levels without interference from vertical mixing.
\end{itemize}

The horizontal slicing approach also entails certain limitations. First, vertical motion components, which may be relevant during intense convective events, are not explicitly modeled. Second, interactions between neighboring altitude levels cannot directly influence the estimated motion at a given height. However, explicitly enforcing vertical coupling would likely further homogenize the estimated motion fields, suppressing subtle differences in motion across different altitudes. Independent motion estimation for individual layers should emphasize subtle inter-layer differences where they exist.

For faster inference, we modify the model's data loading pipeline to treat each altitude slice as an independent sample and process all slices of a single volume in parallel as a batch. This batching strategy is used solely for computational efficiency and does not introduce information exchange between altitude levels.

\subsection{Network Architecture}

\begin{figure*}
    \centering
    \includegraphics[width=\linewidth]{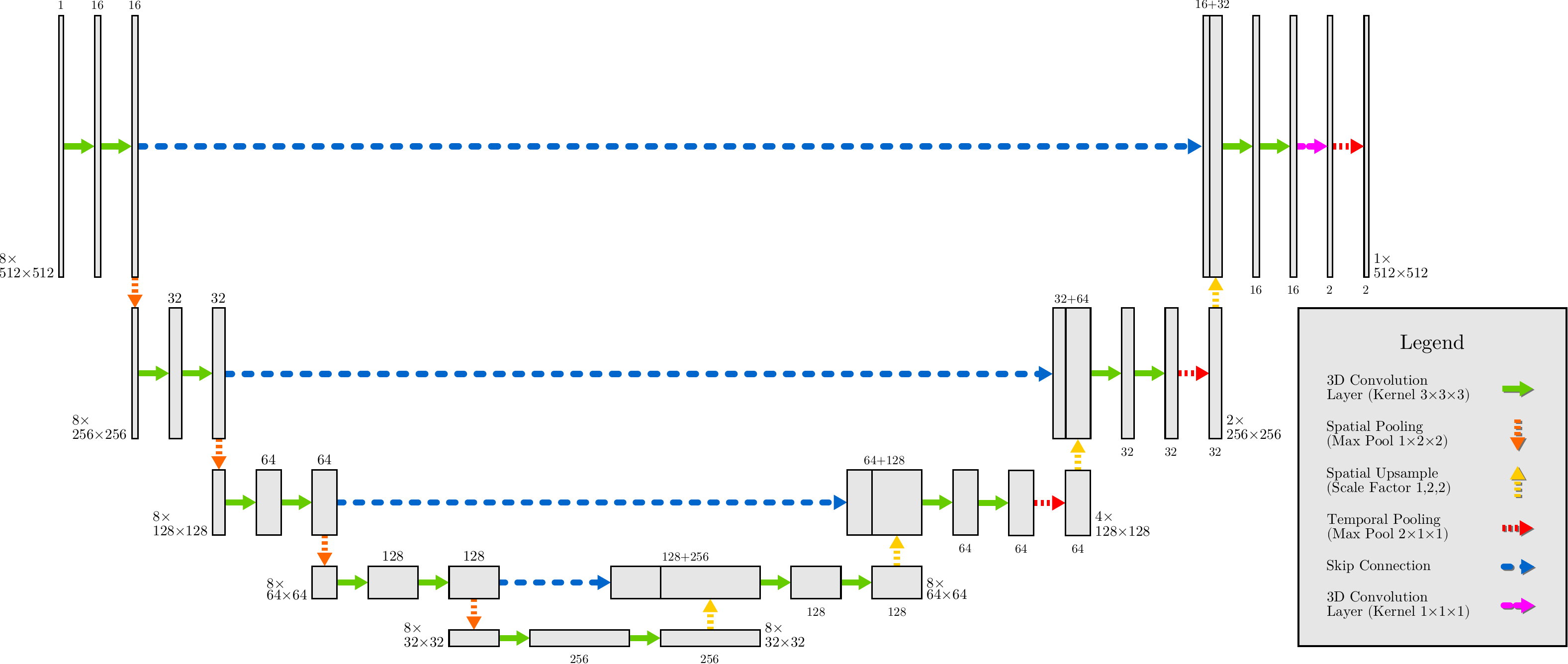}
    \caption{The figure illustrates the 3D U-Net architecture used for horizontal motion field estimation. Feature map blocks indicate tensor dimensions, with the first dimension representing time and the remaining two corresponding to horizontal spatial coordinates. The encoder progressively reduces spatial resolution while increasing channel depth using 3D convolutions and spatial pooling, whereas the decoder restores resolution via upsampling and reduces the number of channels by convolutional processing. Skip connections link corresponding encoder and decoder stages to preserve fine-scale spatial information. The temporal dimension is gradually collapsed in the decoder, yielding a single motion field representing the dominant advection over the input (and near-future) radar sequence. The final layer outputs two channels corresponding to the horizontal motion components.}
    \label{fig:unet}
\end{figure*}

We employ a 3D U-Net architecture (Fig.~\ref{fig:unet}) to estimate horizontal motion fields from spatiotemporal radar volumes. The network follows a standard U-Net encoder–decoder design with skip connections~\cite{ronneberger2015u}, but uses 3D convolutions to explicitly capture temporal dynamics, rather than treating time as independent channels.

In the encoder, horizontal spatial resolution is progressively reduced while feature dimensionality increases, enabling the network to capture increasingly abstract motion patterns across time. The temporal dimension is preserved throughout this stage to retain information about the evolution of precipitation structures.

The decoder restores spatial resolution and integrates multi-scale features via skip connections. During decoding, the temporal dimension is gradually aggregated, resulting in a single representation that describes motion across the input sequence.

The network outputs a two-channel field corresponding to the horizontal components of the estimated motion, representing the dominant advection of precipitation over the input and into the immediate future.

\subsection{Model Training}

Estimating physically consistent horizontal advection fields $\boldsymbol{u}$ from radar data is challenging, as the motion field must capture coherent displacement patterns while remaining spatially smooth and physically plausible. A naïve formulation based on minimizing single-step extrapolation loss:
\begin{equation}
\mathcal{L}_{\text{naïve}} = \mathcal{C}\!\left(\xi^{1}(\Psi_t, \boldsymbol{u}), \hat{\Psi}_{t+1}\right),
\end{equation}
where $\xi^{1}(\Psi_t, \boldsymbol{u})$ denotes the precipitation field $\Psi_t$ advected forward by one time step using motion field $\boldsymbol{u}$, $\mathcal{C}$ is an error criterion, and $\hat{\Psi}_{t+1}$ the ground-truth observation, tends to produce motion fields that overfit individual frame transitions and exhibit unrealistic local divergence.

To address this, we adopt a sequence-consistent training objective, where a single motion field is trained to explain the evolution of precipitation across both observed and future frames. Given an input sequence $\{\Psi_1, \dots, \Psi_n\}$ and future observations $\{\hat{\Psi}_{n+1}, \dots, \hat{\Psi}_{n+m}\}$, the loss is defined as
\begin{equation}
\mathcal{L}_{\text{MF}} = \sum_{t=1}^{n+m-1} 
\mathcal{C}\!\left(\xi^{1}(\Phi_t, \boldsymbol{u}), \Phi_{t+1}\right),
\end{equation}
where $\Phi = \{\Psi_1, \dots, \Psi_n, \hat{\Psi}_{n+1}, \dots, \hat{\Psi}_{n+m}\}$ denotes the concatenated sequence. This encourages temporally consistent motion estimates that remain predictive over the nowcasting horizon.

To further stabilize training, the extrapolation loss is evaluated across multiple spatial scales via average pooling, promoting coherent large-scale motion while retaining sensitivity to finer structures.

In addition, we incorporate a physics-informed regularization term that penalizes violations of the continuity equation and conservation of mass by minimizing the absolute divergence of the motion field,
\begin{equation}
\mathcal{L}_{\text{PI}} = \left| \nabla \cdot \boldsymbol{u} \right|,
\end{equation}
where divergence $\nabla \cdot \boldsymbol{u}$ is approximated using discrete convolutions by Sobel filters.

The final weighted loss combines the data-driven and physics loss:
\begin{equation}
\mathcal{L} = (1 - \beta)\,\mathcal{L}_{\text{MF}} + \beta\,\mathcal{L}_{\text{PI}}, \quad \beta \in (0,1).
\end{equation}

\subsection{Training Data Preprocessing}

The radar observations undergo a series of preprocessing steps designed to ensure physical consistency, numerical stability, and computational efficiency, while preserving the information relevant for motion estimation.

Reflectivity values are converted to rain rate using the Marshall--Palmer relationship~\cite{marshall-palmer} and transformed to a logarithmic scale, with low-intensity values clipped to a fixed minimum.

Each training sample consists of 8 input frames (40 minutes) and 16 future frames (80 minutes), the latter used only in loss computation. All volumes are denoised using the procedure described in Section~\ref{sec:dataset}.

During training, random $512 \times 512$ spatial crops are extracted to match model resolution constraints and increase spatial diversity. Data augmentation is applied via random horizontal and vertical flips. Regions outside radar coverage are masked and excluded from loss computation.

To reduce computational cost, the original 16 altitude levels are pooled into 8 levels using max-pooling of adjacent layers. Given the strong inter-altitude correlation observed in the dataset, it should be safe to omit to reduce computational load.

\section{Evaluation} \label{sec:eval}

We evaluate the proposed volumetric motion field framework in two ways. First, we compare its nowcasting performance against a controlled 2D baseline. Second, we analyze the properties of the estimated volumetric motion fields across the whole dataset, focusing on their vertical consistency and potential added value.

\subsection{Nowcasting Performance of 2D vs.\ 3D Extrapolation}
\label{sec:2D3D}

To assess the benefit of volumetric motion estimation, we compare the proposed model against a baseline with identical architecture trained on 2D CMAX inputs. The only difference lies in the input: a single pooled image versus eight altitude-resolved slices. This ensures that any performance differences arise solely from the use of volumetric motion.

Both models are trained on data up to January~1st,~2022, with the remaining period reserved for evaluation. Simple nowcasts are generated by advecting the last observed field using the estimated motion under the Lagrangian persistence assumption. Outputs of the volumetric model are collapsed to CMAX to enable direct comparison.

Performance is evaluated using continuous and categorical metrics. Mean Error (ME) captures bias, while Mean Absolute Error (MAE) and Mean Squared Error (MSE) measure overall accuracy. Additionally, Precision, Recall, and Equitable Threat Score (ETS) are computed for thresholds of 1, 5, and 10~\si{\mm\per\hour} across the 80-minute forecast horizon.

\subsubsection{Quantitative Evaluation}

We first compare the Mean Absolute Error (MAE) and Mean Squared Error (MSE) of the baseline and volumetric models. Both metrics are expected to behave similarly, with MSE emphasizing larger errors. Both are affected heavily by the double-penalty effect, but they provide insight into error growth with lead time.

\begin{figure}
    \centering
    \includegraphics[width=0.9\columnwidth]{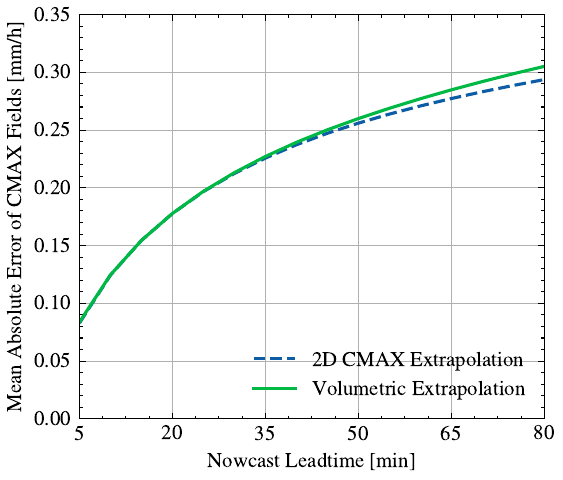}
    \caption{Mean absolute error of extrapolation-based nowcasts as a function of lead time. The blue dashed line corresponds to the 2D motion field model, while the solid green line represents the volumetric motion field model operating on altitude-wise radar slices.}
    \label{fig:2Dvs3DMAE}
\end{figure}

\begin{figure}
    \centering
    \includegraphics[width=0.9\columnwidth]{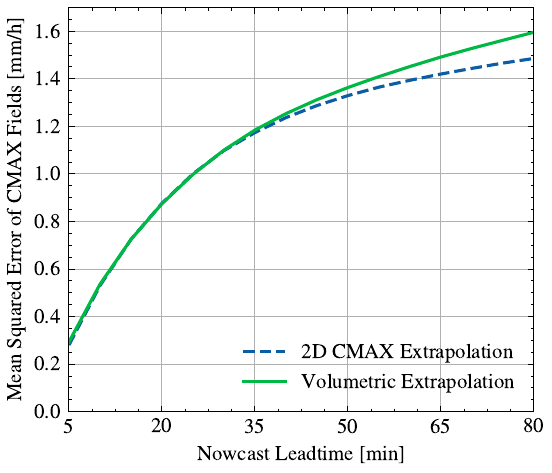}
    \caption{Mean squared error of extrapolation-based nowcasts as a function of lead time. The blue dashed line corresponds to the 2D motion field model, while the solid green line represents the volumetric motion field model operating on altitude-wise radar slices.}
    \label{fig:2Dvs3DMSE}
\end{figure}

Figure~\ref{fig:2Dvs3DMAE} shows that MAE is nearly identical for both models up to approximately 30~minutes. Beyond this point, the volumetric model exhibits slightly higher errors. The same pattern is observed for MSE in Fig.~\ref{fig:2Dvs3DMSE}.

\begin{figure}
    \centering
    \includegraphics[width=0.9\columnwidth]{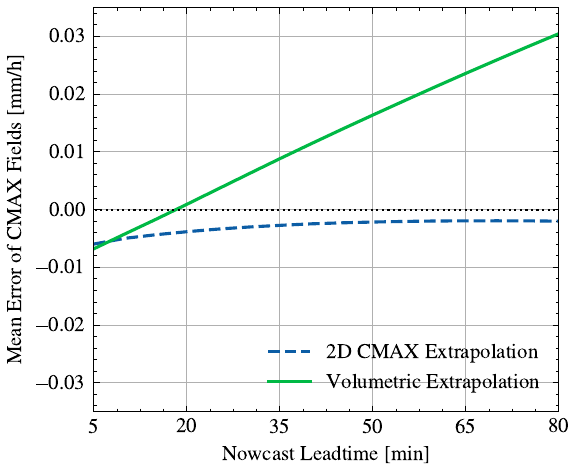}
    \caption{Mean error (bias) of extrapolation-based nowcasts as a function of lead time. The blue dashed line corresponds to the 2D motion field model, while the solid green line represents the volumetric motion field model operating on altitude-wise radar slices. The black dotted line indicates zero bias.}
    \label{fig:2Dvs3DME}
\end{figure}

The Mean Error (ME) in Fig.~\ref{fig:2Dvs3DME} reveals differing bias behavior. Both models start by slightly underestimating the precipitation at short lead times. The 2D model remains near-neutral, whereas the volumetric model develops an increasing positive bias.

Overall, continuous metrics suggest slightly worse performance of the volumetric model at longer lead times. However, due to their sensitivity to displacement errors, we also compute categorical metrics.

\begin{figure*}[t]
\centering

\begin{tabular}{c p{0.29\textwidth} p{0.29\textwidth} p{0.29\textwidth}}
    & \centering\arraybackslash{1 \si{\mm\per\hour}}
    & \centering\arraybackslash{5 \si{\mm\per\hour}}
    & \centering\arraybackslash{10 \si{\mm\per\hour}} \\[3ex]

    \rotatebox{90}{Precision} &
    \multicolumn{3}{c}{
        $\vcenter{\hbox{\includegraphics[width=0.87\linewidth]{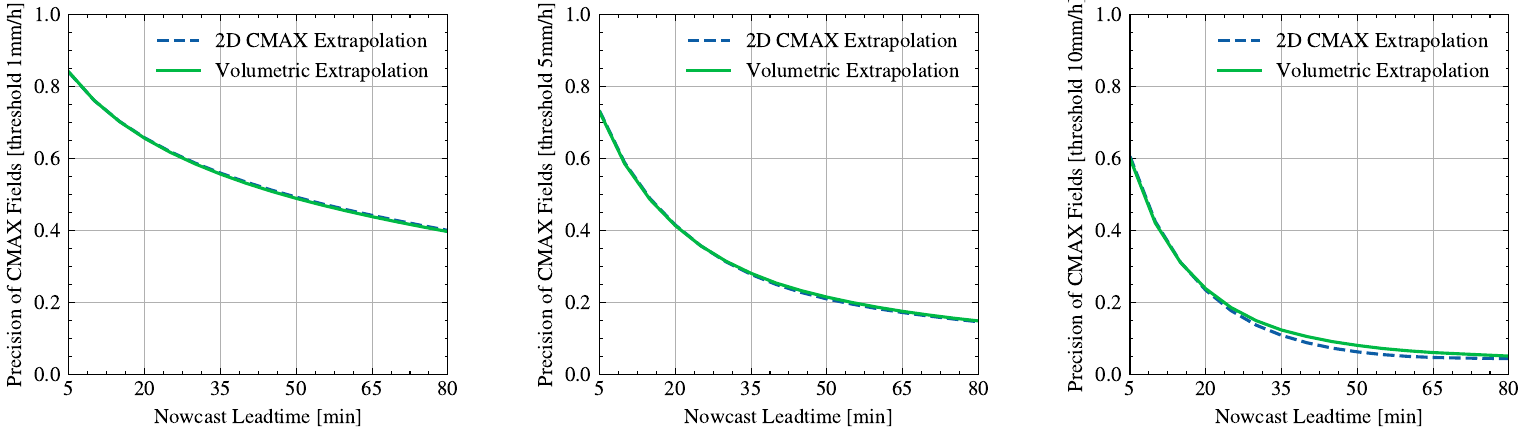}}}$
    } \\

    \rotatebox{90}{Recall} &
    \multicolumn{3}{c}{
        $\vcenter{\hbox{\includegraphics[width=0.87\linewidth]{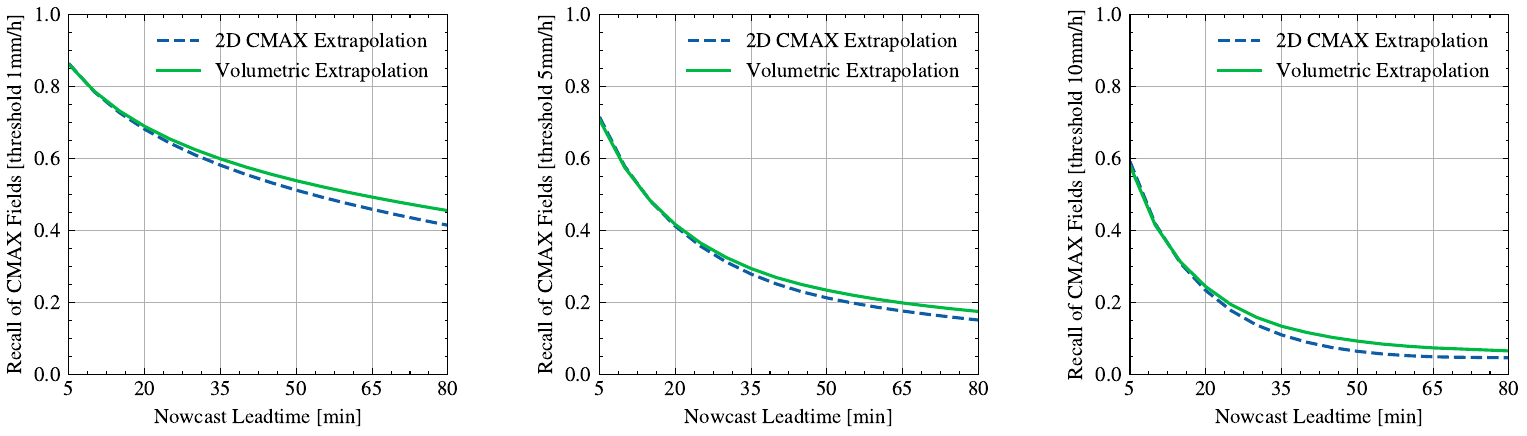}}}$
    } \\

    \rotatebox{90}{ETS} &
    \multicolumn{3}{c}{
        $\vcenter{\hbox{\includegraphics[width=0.87\linewidth]{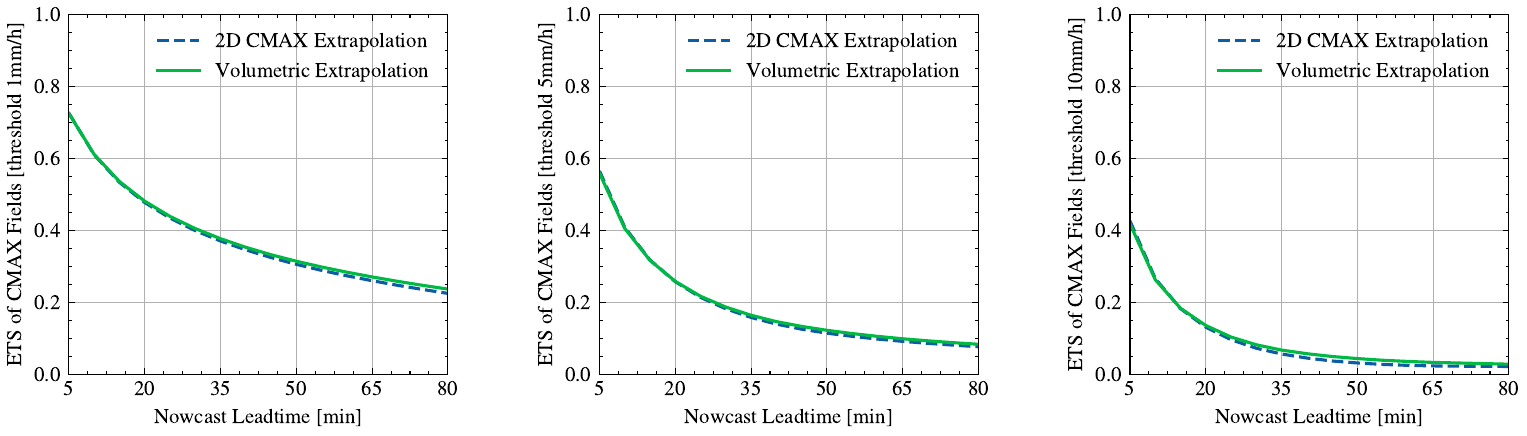}}}$
    }
\end{tabular}

\caption{The figure shows categorical verification metrics -- Precision, Recall, and Equitable Threat Score (ETS) -- as functions of forecast lead time for the baseline 2D motion-field extrapolation model and the proposed volumetric extrapolation model. Results are shown for precipitation intensity thresholds of 1, 5, and 10~\si{\mm\per\hour}. Columns correspond to a fixed binarization threshold, while rows compare the two models using the same metric across the selected thresholds.}
\label{fig:2Dvs3DThresholded}
\end{figure*}

Figure~\ref{fig:2Dvs3DThresholded} shows Precision, Recall, and ETS for thresholds of 1, 5, and 10~\si{\mm\per\hour}. The volumetric model consistently matches or improves upon the baseline. The largest gains occur in Recall beyond 30~minutes, while Precision remains comparable or slightly higher at higher thresholds. ETS shows a small but consistent improvement.

The increase in Recall at all thresholds is notable, indicating better detection of precipitation at longer lead times. However, together with the growing positive bias, this suggests a tendency toward overestimation.

\begin{figure}
    \centering
    \includegraphics[width=0.9\columnwidth]{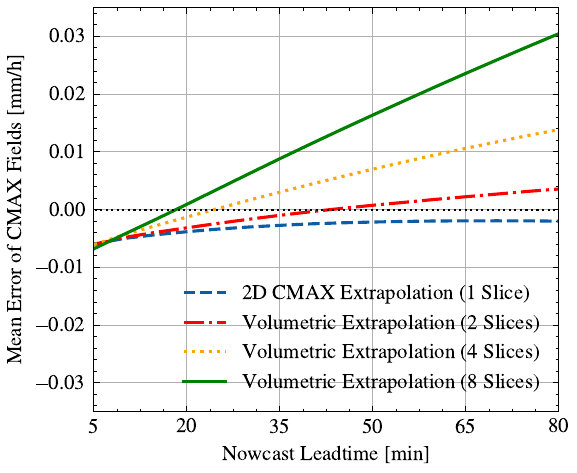}
    \caption{Mean error (bias) of extrapolation-based nowcasts as a function of lead time. In addition to the baseline 2D CMAX model and the 8-level volumetric model shown in Fig.~\ref{fig:2Dvs3DME}, we include intermediate configurations operating on data pooled to 2 and 4 vertical levels. The black dotted line indicates zero bias.}
    \label{fig:2Dvs3DME_all}
\end{figure}

To investigate this, we train additional volumetric models with 2 and 4 altitude levels. Figure~\ref{fig:2Dvs3DME_all} shows that bias increases systematically with the number of altitude levels, indicating that the effect is structural.

This behavior motivates a qualitative analysis in the following subsection.

\subsubsection{Qualitative Evaluation}

To complement the quantitative results, we analyze a single severe precipitation event from the test set, selected as the most intense case in terms of reflectivity and spatial extent. Such an extreme event provides a suitable setting to assess the behavior of the volumetric motion field estimator under conditions where vertical variability is expected.

The event follows a heatwave and is associated with a convective outbreak along a cold front over western Slovakia, producing intense rainfall, strong winds, and organized storm structures. These conditions suggest potential benefits from altitude-wise motion estimation.

\begin{figure*}
    \centering
    \includegraphics[width=0.328\textwidth]{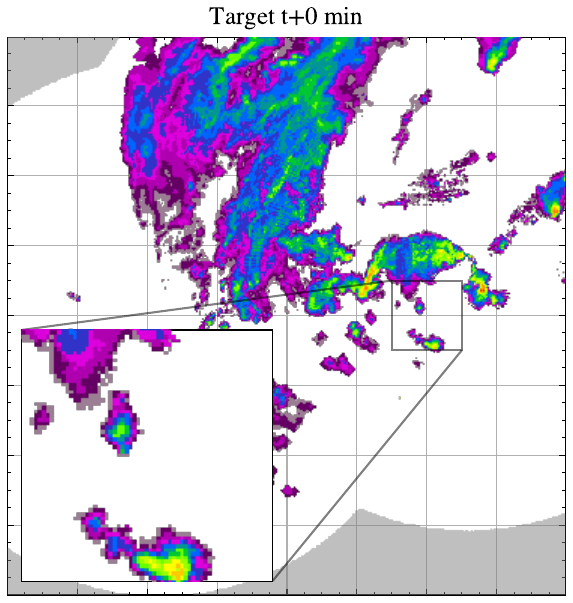}
    \includegraphics[width=0.328\textwidth]{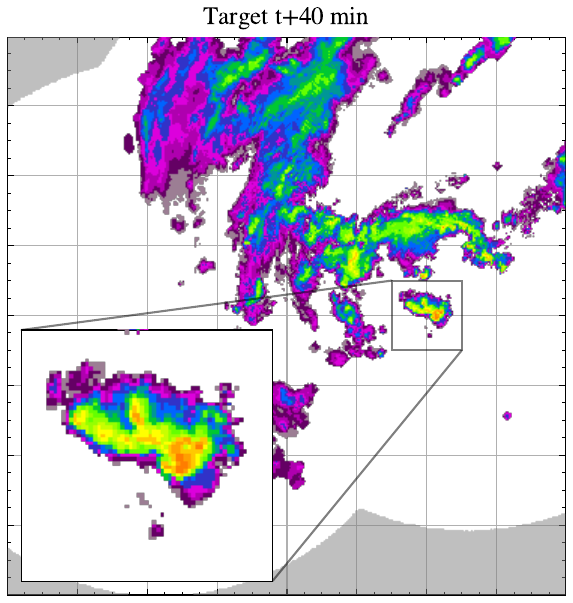}
    \includegraphics[width=0.328\textwidth]{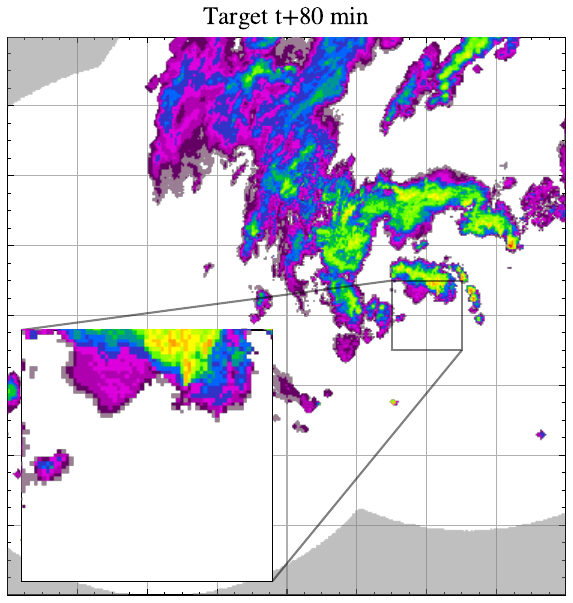}

    \includegraphics[width=0.328\textwidth]{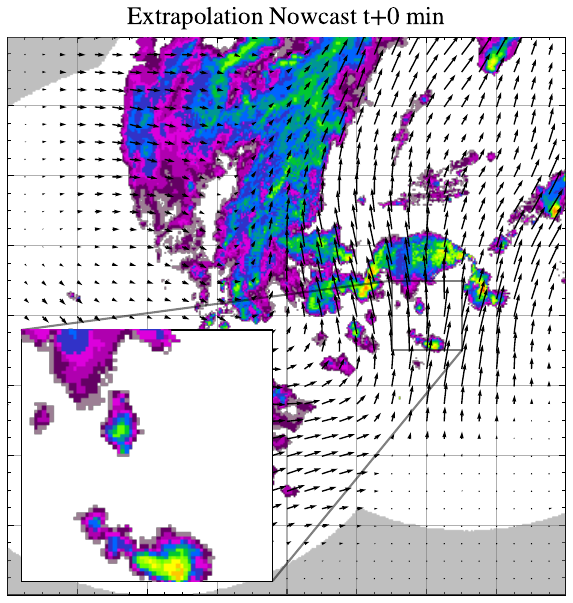}
    \includegraphics[width=0.328\textwidth]{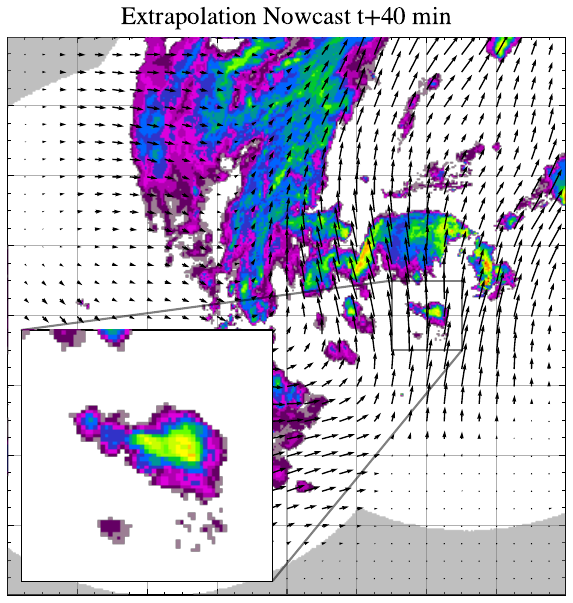}
    \includegraphics[width=0.328\textwidth]{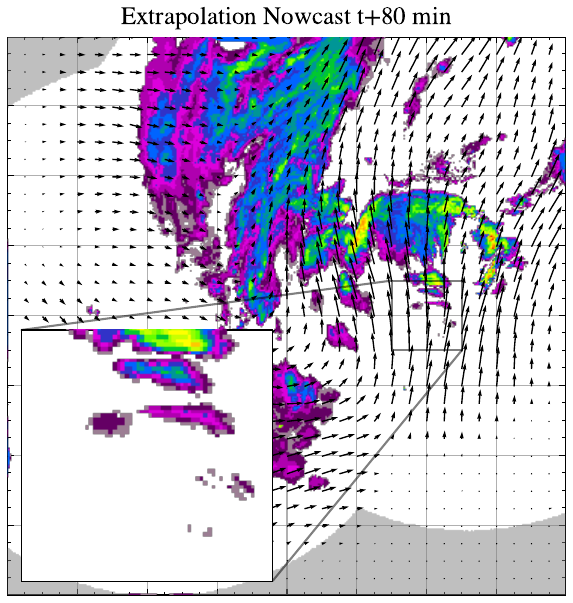}

    \caption{Vertically pooled CMAX representations of a severe precipitation event over Slovakia on July 5th 2022. The top row shows three consecutive radar observations, with the first image (t+0~min) corresponding to 15:15~UTC and the following two images taken at 40-minute intervals. The bottom row shows the corresponding extrapolations obtained by advecting the final observation using the volumetric motion field estimated by the proposed 3D Motion Field U-Net. The vertically averaged 3D motion field is overlaid for reference. Zoomed regions highlight a non-physical cell-splitting artifact produced by the volumetric approach. The animated figure is available online at: \url{https://pavlikp.github.io/3DMF/\#figure-12}}
    \label{fig:qualitative_artifacts}
\end{figure*}

Due to visualization constraints, results are shown as vertically pooled CMAX fields, while interpretation is based on the full 3D volumes. Figure~\ref{fig:qualitative_artifacts} presents an 80-minute volumetric nowcast of the event.

Although magnitudes of estimated motion fields vary across altitudes, their directions remain largely aligned, indicating strong vertical coherence and limited benefit from independent altitude-wise estimation in this case.

The zoomed regions reveal the source of the increasing positive bias. While the model captures the overall northward displacement accurately, the forecast exhibits a splitting of a single cell into multiple structures at longer lead times. This artifact arises from small discrepancies between motion fields at different altitudes: higher layers advect more slowly, producing apparent splitting when collapsed into a CMAX representation. The ground truth, in contrast, shows a single coherent and growing cell.

The non-physical cell-splitting -- such as the one shown here -- increases the spatial extent of predicted precipitation, contributing to overestimation and explaining the observed improvement in Recall. However, it does not reflect realistic storm evolution. The artifact results from independently estimated motion fields across altitudes combined with evaluation on vertically pooled outputs, highlighting a key limitation of the current approach and motivating the need for vertically consistent motion estimation.

\subsection{Motion Field Correlation Analysis} \label{sec:MDcorr}

Although the proposed volumetric model does not significantly improve overall nowcasting performance, the estimated three-dimensional motion fields remain informative. There might still be cases when volumetric motion fields can bring considerable benefits. We therefore analyze their internal structure, focusing on the similarity of motion estimates across altitude levels. Strong inter-layer correlation would indicate redundant vertical information, while lower correlation may reveal situations where altitude-dependent motion provides added value.

Motion fields are computed for all samples, including both training and test data, as the goal is to characterize the estimated fields rather than assess generalization. Based on this analysis, we identify cases with low inter-altitude correlation for further qualitative inspection.

\subsubsection{Quantitative Evaluation}

To quantify vertical coherence, we compute pairwise Pearson correlation coefficients between horizontal motion fields at different altitude levels. Correlations are evaluated only over precipitating regions, using a binary mask derived from the corresponding input slices to exclude non-informative areas.

\begin{figure}
    \centering
    \includegraphics[width=\columnwidth]{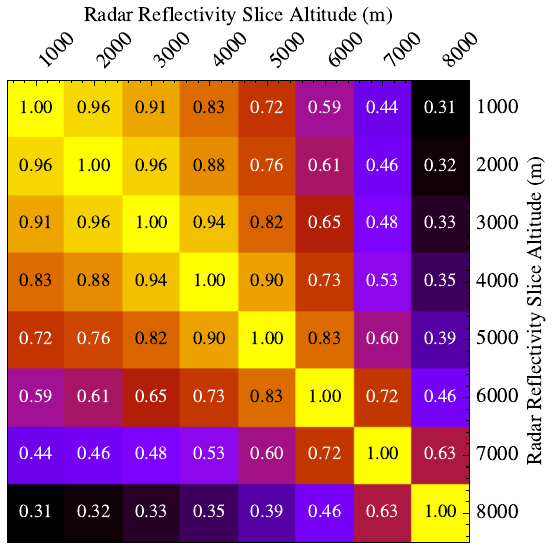}
    \caption{Mean Pearson correlation coefficients between horizontal motion fields estimated at different altitude levels. Correlations are computed only over precipitating regions and averaged over the full dataset.
    Note: Each horizontal layer consists of a max-pooled pair of neighboring radar volume slices (altitude 1000 \si{\meter} represents pooled layers at 500 and 1000 \si{\meter} and so on).}
    \label{fig:mf_corr}
\end{figure}

Figure~\ref{fig:mf_corr} shows the mean correlation matrix for the model operating on 8 altitude levels. Adjacent layers exhibit very strong correlations (typically $>0.9$), with values decreasing gradually as vertical separation increases. Even for the most distant layers, correlations remain moderate (above $0.3$).

Overall, the motion fields are highly vertically coherent, with only gradual decorrelation with height. This suggests that dominant advection patterns are largely shared across altitudes, explaining the limited gains of volumetric nowcasting we observed.

To assess whether more favorable cases exist, we search for samples with unusually low inter-layer correlation. Prior work~\cite{chung2025investigating} has reported events with substantial vertical differences in motion, and we investigate whether similar situations occur in our dataset and whether they offer practical benefits for altitude-wise motion estimation.

\subsubsection{Qualitative Evaluation}

We look for samples combining high precipitation coverage with low inter-altitude motion field correlation, as these are the most likely to exhibit meaningful vertical differences in horizontal advection.

\begin{figure*}
    \centering
    \includegraphics[width=\linewidth]{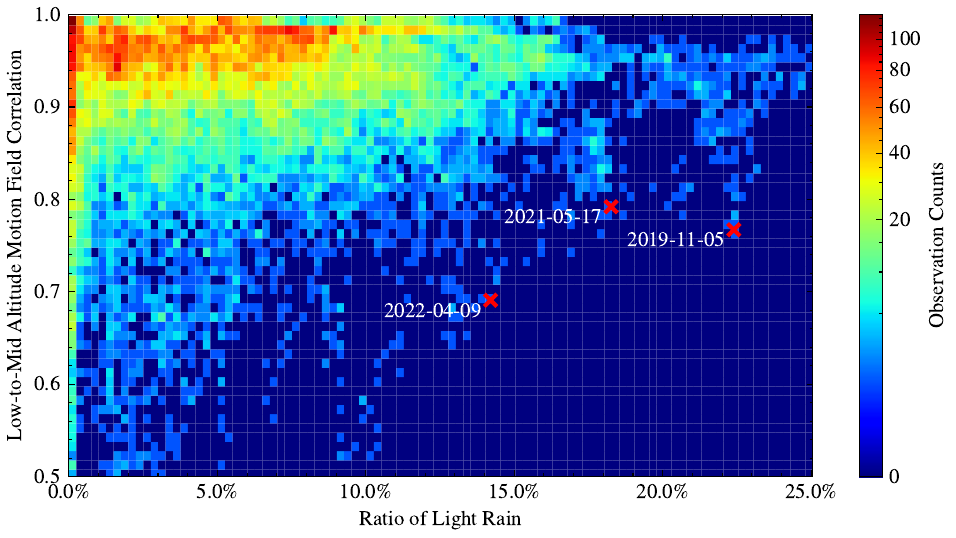}
    \caption{Two-dimensional histogram of dataset samples as a function of precipitation coverage and inter-altitude motion field correlation. The horizontal axis shows the ratio of CMAX-pooled pixels exceeding 20~\si{\dB Z}, while the vertical axis shows the Pearson correlation between horizontal motion fields estimated at the lowest altitude layer (500–1000~\si{\meter} above ground level) and a mid-level layer (2500–3000~\si{\meter} above ground level). Red hues indicate the highest density of samples, dark blue no such samples. The three crosses denote events selected for qualitative evaluation.}
    \label{fig:outliers}
\end{figure*}

To identify such cases, we construct a two-dimensional density histogram over the full dataset (Fig.~\ref{fig:outliers}), relating precipitation coverage (ratio of CMAX pixels exceeding 20~\si{\dB Z}) to inter-layer Pearson correlation between low (500–1000~\si{\meter}) and mid-level (2500–3000~\si{\meter}) motion fields. Regions with high coverage and low correlation are sparsely populated, indicating that such conditions are rare.

We select three representative outlier events using a joint ranking based on high precipitation coverage and low correlation, excluding temporally adjacent samples from the same event.

\begin{figure*}
    \centering

    \includegraphics[width=0.7\linewidth]{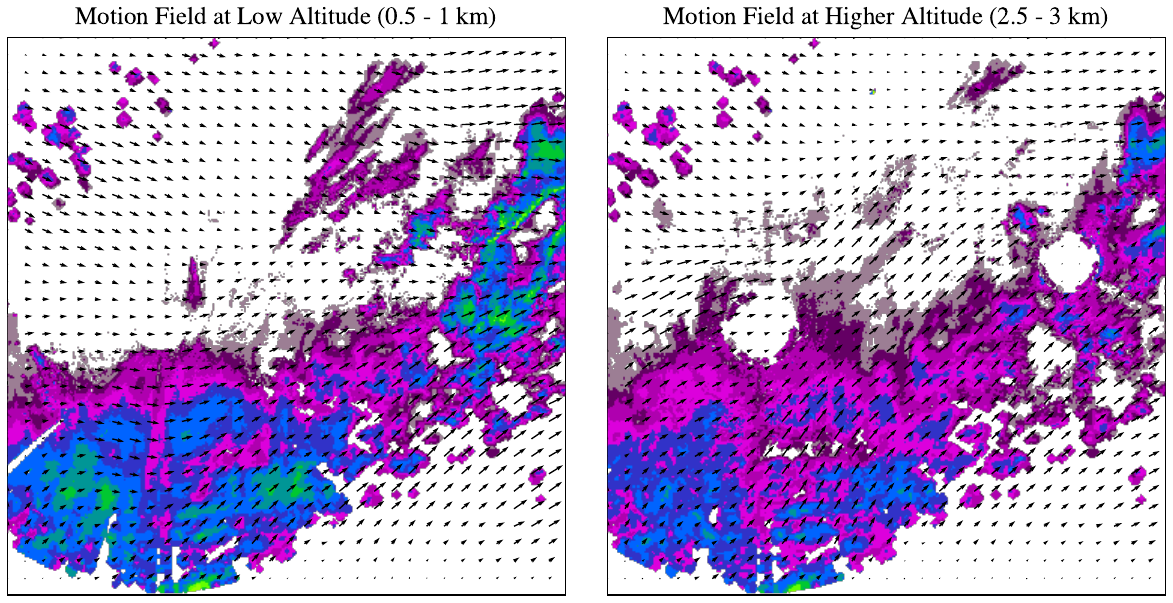}
    \par\small\centering (a) 2022-04-09 09:55
    \vspace{0.5cm}

    \includegraphics[width=0.7\linewidth]{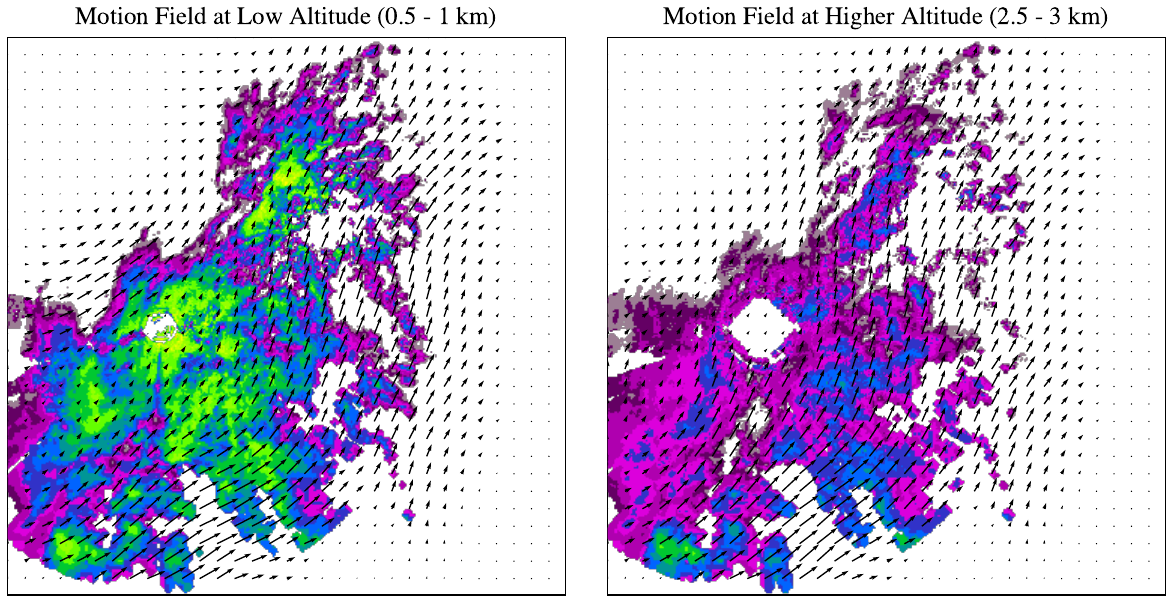}
    \par\small\centering (b) 2021-05-17 05:50
    \vspace{0.5cm}

    \includegraphics[width=0.7\linewidth]{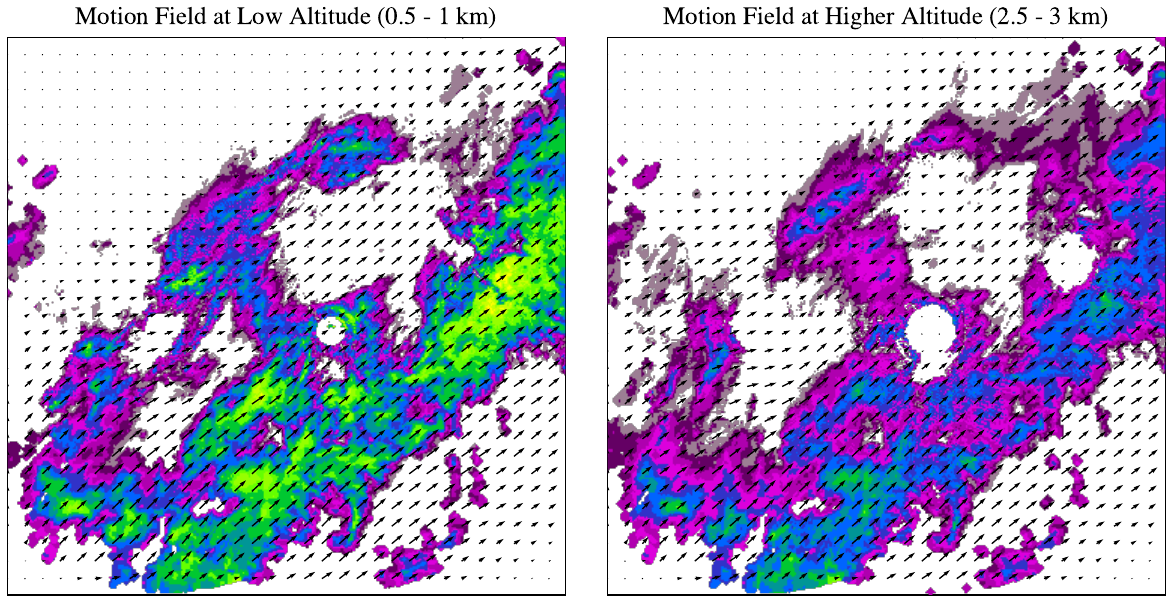}
    \par\small\centering (c) 2019-11-05 20:30

    \caption{Estimated horizontal motion fields and corresponding precipitation fields at the lowest altitude layer (500–1000~\si{\meter} above ground level; left) and a mid-level layer (2500–3000~\si{\meter} above ground level; right) for three events selected for their low inter-altitude motion correlation and high precipitation coverage. The animated figures are available online at: \url{https://pavlikp.github.io/3DMF/\#figure-16}}
    \label{fig:qualitative}
\end{figure*}

For each case, we visualize motion fields at low (500 to 1000~\si{\meter} above ground level) and mid (2500 to 3000~\si{\meter} above ground level) altitudes alongside the corresponding precipitation fields (Fig.~\ref{fig:qualitative}).

Across all cases, motion fields at low and mid altitudes remain visually similar. The largest discrepancy appears in the lower-left region of event~(a), where motion shifts from eastward to north-eastward. However, this region coincides with data artifacts (e.g., ground clutter or occlusion), suggesting that the difference is not physically meaningful.

In the remaining cases, motion fields are strongly aligned in direction, with only minor magnitude differences, primarily in data-sparse regions. Despite explicitly selecting events with as low inter-altitude correlation as possible (for a given precipitation coverage), no clear example emerges where altitude-resolved motion provides a substantial advantage over two-dimensional estimation.

These findings confirm that vertical variability in horizontal precipitation motion is generally weak in this dataset. This is consistent with the strong inter-altitude correlations observed earlier and explains the limited benefits in the nowcasting performance of the volumetric approach.

\section{Conclusion}

We proposed a 3D Motion Field U-Net for independently estimating altitude-wise horizontal advection of precipitation from volumetric radar sequences, with the goal of improving extrapolation-based nowcasting.

The proposed model was compared to an architecturally identical baseline operating on vertically pooled 2D composites. While the volumetric model improved categorical metrics, particularly Recall at longer lead times, it underperformed in continuous error metrics. Qualitative analysis also showed that the gains likely result from non-physical cell-splitting artifacts caused by independently estimated motion fields across altitudes and vertically pooled nowcasts, rather than improved representation of storm dynamics.

The computational efficiency of the framework enabled dataset-scale analysis of the estimated motion fields. This revealed strong inter-altitude correlations throughout the radar volumes, with gradual decorrelation at larger vertical separations. Even in carefully selected cases with high precipitation coverage and relatively low inter-altitude correlation, motion fields remained largely aligned, exhibiting almost no vertical variability in horizontal advection.

These findings suggest that, for the Slovak radar dataset, altitude-wise motion estimation does not provide a significant systematic advantage over traditional 2D approaches. Vertical differences in horizontal motion appear too weak to be effectively exploited.

The study has several limitations. It is based on a single regional dataset and assumes pure Lagrangian persistence without modeling intensity evolution. Additionally, motion fields are estimated independently across altitudes, without enforcing vertical consistency, contributing to observed artifacts. Future work should explore vertically coupled or physically constrained motion estimation.

Overall, the results indicate that the added complexity of volumetric motion modeling may not be justified in regions dominated by vertically coherent precipitation systems. However, the proposed framework remains relevant for climates with more frequent vertical wind shear and heterogeneous storm structure. More broadly, this work highlights the importance of critically assessing the practical benefits of three-dimensional representations in data-driven precipitation nowcasting and spatiotemporal motion estimation more generally.

\section*{Code Repository}
Training and evaluation source code, along with a link to download the used dataset are available at \url{https://github.com/kinit-sk/volumetric-motion-fields}.

\section*{Funding}

This work was partially funded by the European Union, under the project LorAI - Low Resource Artificial Intelligence, GA No. \href{https://doi.org/10.3030/101136646}{101136646}; and  by V4Grid, an Interreg Central Europe Programme project co-funded by the European Union, project No. CE0200803. The research was performed in cooperation with Softec, Ltd. and Slovak Hydrometeorological Institute.

Research results were obtained using the computational resources procured in the national project National competence centre for high performance computing (project code: 311070AKF2) funded by European Regional Development Fund, EU Structural Funds Informatization of society, Operational Program Integrated Infrastructure.

\section*{Conflict of interest}

The authors have no competing interests to declare that are relevant to the content of this article.

\section*{Declaration of generative AI and AI-assisted technologies in the manuscript preparation process}

During the preparation of this work the authors used ChatGPT to assist with language editing and manuscript refinement. After using this service, the authors reviewed and edited the content as needed and take full responsibility for the content of the published article.

\printcredits

\bibliographystyle{cas-model2-names}

\bibliography{cas-refs}

@article{marshall-palmer,
    author = "J. S. Marshall and W. Mc K. Palmer",
    title = {{The Distribution of Raindrops with Size}},
    journal = "Journal of the Atmospheric Sciences",
    year = "1948",
    volume = "5",
    number = "4",
    pages = "165--166",
    url = {https://doi.org/10.1175/1520-0469(1948)005<0165:TDORWS>2.0.CO;2}
}

@ARTICLE{lcnn,
  author={Ritvanen, Jenna and Harnist, Bent and Aldana, Miguel and Mäkinen, Terhi and Pulkkinen, Seppo},
  journal={IEEE Journal of Selected Topics in Applied Earth Observations and Remote Sensing}, 
  title={{Advection-Free Convolutional Neural Network for Convective Rainfall Nowcasting}}, 
  year={2023},
  volume={16},
  number={},
  pages={1654-1667},
  doi={10.1109/JSTARS.2023.3238016}}

@article{zhang2023skilful,
  title={{Skilful nowcasting of extreme precipitation with NowcastNet}},
  author={Zhang, Yuchen and Long, Mingsheng and Chen, Kaiyuan and Xing, Lanxiang and Jin, Ronghua and Jordan, Michael I and Wang, Jianmin},
  journal={Nature},
  volume={619},
  number={7970},
  pages={526--532},
  year={2023},
  publisher={Nature Publishing Group UK London}
}

@article{Keil2009,
  author = "Christian  Keil and George C.  Craig",
  title = {{{"A Displacement and Amplitude Score Employing an Optical Flow Technique"}}},
  journal = "Weather and Forecasting",
  year = "2009",
  publisher = "American Meteorological Society",
  address = "Boston MA, USA",
  volume = "24",
  number = "5",
  doi = "10.1175/2009WAF2222247.1",
  pages=      "1297 - 1308",
  url = "https://journals.ametsoc.org/view/journals/wefo/24/5/2009waf2222247_1.xml"
}

@article{ayzel2020rainnet,
  title={{RainNet v1. 0: a convolutional neural network for radar-based precipitation nowcasting}},
  author={Ayzel, Georgy and Scheffer, Tobias and Heistermann, Maik},
  journal={Geoscientific Model Development},
  volume={13},
  number={6},
  pages={2631--2644},
  year={2020},
  publisher={Copernicus GmbH}
}

@article{shi2015convolutional,
  title={{Convolutional LSTM network: A machine learning approach for precipitation nowcasting}},
  author={Shi, Xingjian and Chen, Zhourong and Wang, Hao and Yeung, Dit-Yan and Wong, Wai-Kin and Woo, Wang-chun},
  journal={Advances in neural information processing systems},
  volume={28},
  year={2015}
}

@article{shi2017deep,
  title={{Deep learning for precipitation nowcasting: A benchmark and a new model}},
  author={Shi, Xingjian and Gao, Zhihan and Lausen, Leonard and Wang, Hao and Yeung, Dit-Yan and Wong, Wai-kin and Woo, Wang-chun},
  journal={Advances in neural information processing systems},
  volume={30},
  year={2017}
}

@Book{WMO17,
  author =       "{WMO}",
  title =        {{Guidelines for Nowcasting Techniques}},
  publisher =    "World Meteorological Organization",
  year =         "2017",
  address =      "7 bis, avenue de la Paix, P.O. Box 2300, CH-1211 Geneva 2, Switzerland",
  edition =      "",
  editor =       "",
  volume =       "",
  number =       "978-92-63-11198-2",
}

@inproceedings{lucas1981iterative,
  title={{An iterative image registration technique with an application to stereo vision}},
  author={Lucas, Bruce D and Kanade, Takeo},
  booktitle={IJCAI'81: 7th international joint conference on Artificial intelligence},
  volume={2},
  pages={674--679},
  year={1981}
}

@article{pulkkinen2021lagrangian,
  title={Lagrangian integro-difference equation model for precipitation nowcasting},
  author={Pulkkinen, Seppo and Chandrasekar, V and Niemi, Tero},
  journal={Journal of Atmospheric and Oceanic Technology},
  volume={38},
  number={12},
  pages={2125--2145},
  year={2021}
}

@article{bowler2006steps,
  title={{STEPS: A probabilistic precipitation forecasting scheme which merges an extrapolation nowcast with downscaled NWP}},
  author={Bowler, Neill E and Pierce, Clive E and Seed, Alan W},
  journal={Quarterly Journal of the Royal Meteorological Society: A journal of the atmospheric sciences, applied meteorology and physical oceanography},
  volume={132},
  number={620},
  pages={2127--2155},
  year={2006},
  publisher={Wiley Online Library}
}

@article{
Zhang2024,
author = {Wenxia Zhang  and Tianjun Zhou  and Peili Wu},
title = {{Anthropogenic amplification of precipitation variability over the past century}},
journal = {Science},
volume = {385},
number = {6707},
pages = {427-432},
year = {2024},
doi = {10.1126/science.adp0212},
URL = {https://www.science.org/doi/abs/10.1126/science.adp0212},
eprint = {https://www.science.org/doi/pdf/10.1126/science.adp0212}
}

@article{NOVAK2009328,
title = {{Quantitative precipitation forecast using radar echo extrapolation}},
journal = {Atmospheric Research},
volume = {93},
number = {1},
pages = {328-334},
year = {2009},
note = {4th European Conference on Severe Storms},
issn = {0169-8095},
doi = {https://doi.org/10.1016/j.atmosres.2008.10.014},
url = {https://www.sciencedirect.com/science/article/pii/S0169809508003025},
author = {Petr Novák and Lucie Březková and Petr Frolík}
}

@article{wang2017predrnn,
  title={{PredRNN: Recurrent neural networks for predictive learning using spatiotemporal LSTMs}},
  author={Wang, Yunbo and Long, Mingsheng and Wang, Jianmin and Gao, Zhifeng and Yu, Philip S},
  journal={Advances in neural information processing systems},
  volume={30},
  year={2017}
}

@article{pavlik2025fully,
  title={{Fully differentiable Lagrangian convolutional neural network for physics-informed precipitation nowcasting}},
  author={Pavl{\'\i}k, Peter and V{\'y}boh, Martin and Ezzeddine, Anna Bou and Rozinajov{\'a}, Viera},
  journal={Applied Computing and Geosciences},
  pages={100296},
  year={2025},
  publisher={Elsevier}
}

@article{sun2022three,
  title={{Three-dimensional gridded radar echo extrapolation for convective storm nowcasting based on 3D-ConvLSTM model}},
  author={Sun, Nengli and Zhou, Zeming and Li, Qian and Jing, Jinrui},
  journal={Remote Sensing},
  volume={14},
  number={17},
  pages={4256},
  year={2022},
  publisher={MDPI}
}

@article{KIM2021105774,
title = {{Improving precipitation nowcasting using a three-dimensional convolutional neural network model from Multi Parameter Phased Array Weather Radar observations}},
journal = {Atmospheric Research},
volume = {262},
pages = {105774},
year = {2021},
issn = {0169-8095},
doi = {https://doi.org/10.1016/j.atmosres.2021.105774},
url = {https://www.sciencedirect.com/science/article/pii/S0169809521003306},
author = {Dong-Kyun Kim and Taku Suezawa and Tomoaki Mega and Hiroshi Kikuchi and Eiichi Yoshikawa and Philippe Baron and Tomoo Ushio}
}

@inproceedings{pavlik2022radar,
  title={{Radar-Based Volumetric Precipitation Nowcasting: A 3D Convolutional Neural Network with U-Net Architecture.}},
  author={Pavl{\'\i}k, Peter and Rozinajov{\'a}, Viera and Ezzeddine, Anna Bou},
  booktitle={CDCEO@ IJCAI},
  pages={65--72},
  year={2022}
}

@article{chen2025nowcast3d,
  title={{Nowcast3D: Reliable precipitation nowcasting via gray-box learning}},
  author={Chen, Huaguan and Han, Wei and Sun, Haofei and Lin, Ning and Song, Xingtao and Yang, Yunfan and Tian, Jie and Liu, Yang and Wen, Ji-Rong and Zhang, Xiaoye and others},
  journal={arXiv preprint arXiv:2511.04659},
  year={2025}
}

@article{chung2025investigating,
  title={{Investigating the spatiotemporal characteristics of motion fields using three-dimensional radar echoes to construct an ensemble nowcasting system}},
  author={Chung, Kao-Shen and Hsu, Yu-Chiao and Tsou, Yi-Hao and Lin, Hsin-Hung},
  journal={Quarterly Journal of the Royal Meteorological Society},
  volume={151},
  number={768},
  pages={e4935},
  year={2025},
  publisher={Wiley Online Library}
}

@article {SPROG,
      author = "A. W. Seed",
      title = {{"A Dynamic and Spatial Scaling Approach to Advection Forecasting"}},
      journal = "Journal of Applied Meteorology",
      year = "2003",
      publisher = "American Meteorological Society",
      address = "Boston MA, USA",
      volume = "42",
      number = "3",
      doi = "10.1175/1520-0450(2003)042<0381:ADASSA>2.0.CO;2",
      pages=      "381 - 388",
      url = "https://journals.ametsoc.org/view/journals/apme/42/3/1520-0450_2003_042_0381_adassa_2.0.co_2.xml"
}

@ARTICLE{pulkinnen2020anvil,
  author={Pulkkinen, Seppo and Chandrasekar, V. and von Lerber, Annakaisa and Harri, Ari-Matti},
  journal={IEEE Transactions on Geoscience and Remote Sensing}, 
  title={{Nowcasting of Convective Rainfall Using Volumetric Radar Observations}}, 
  year={2020},
  volume={58},
  number={11},
  pages={7845-7859},
  doi={10.1109/TGRS.2020.2984594}}

@article{vido2024thunderstorm,
  title={{Thunderstorm climatology of Slovakia between 1984--2023}},
  author={Vido, Jaroslav and Bors{\'a}nyi, Peter and Nalevankov{\'a}, Paul{\'\i}na and Hanzelov{\'a}, Miriam and Ku{\v{c}}era, Ji{\v{r}}{\'\i} and {\v{S}}kvarenina, Jaroslav},
  journal={Theoretical and Applied Climatology},
  volume={155},
  number={9},
  pages={8651--8679},
  year={2024},
  publisher={Springer}
}

@article{nagasato2021extension,
  title={{Extension of convolutional neural network along temporal and vertical directions for precipitation downscaling}},
  author={Nagasato, Takeyoshi and Ishida, Kei and Ercan, Ali and Tu, Tongbi and Kiyama, Masato and Amagasaki, Motoki and Yokoo, Kazuki},
  journal={arXiv preprint arXiv:2112.06571},
  year={2021}
}

@incollection{pavlik2025echo,
  title={{Do Echo Top Heights Improve Deep Learning Rainfall Nowcasts? A Case Study in the Netherlands}},
  author={Pavl{\'\i}k, Peter and Schleiss, Marc and Ezzeddine, Anna Bou and Rozinajov{\'a}, Viera},
  booktitle={Transactions on Large-Scale Data-and Knowledge-Centered Systems LVIII},
  pages={66--92},
  year={2025},
  address={Heidelberg},
  publisher={Springer}
}

@article{ye2022msstnet,
  title={{MSSTNet: A multi-scale spatiotemporal prediction neural network for precipitation nowcasting}},
  author={Ye, Yuankang and Gao, Feng and Cheng, Wei and Liu, Chang and Zhang, Shaoqing},
  journal={Remote Sensing},
  volume={15},
  number={1},
  pages={137},
  year={2022},
  publisher={MDPI}
}

@inproceedings{ronneberger2015u,
  title={U-net: Convolutional networks for biomedical image segmentation},
  author={Ronneberger, Olaf and Fischer, Philipp and Brox, Thomas},
  booktitle={International Conference on Medical image computing and computer-assisted intervention},
  pages={234--241},
  year={2015},
  organization={Springer}
}



\end{document}